\documentclass[twocolumn]{article}
\usepackage{usenix}
\usepackage{amsmath}
\usepackage{amsthm}
\usepackage{amsfonts}
\usepackage{graphicx}
\usepackage{dsfont}
\usepackage{caption}
\usepackage{subcaption}
\usepackage{booktabs}
\usepackage{float}
\begin{document}
\title{Poisoning the Unlabeled Dataset of Semi-Supervised Learning}
\author{Nicholas Carlini \\ Google}
\date{}

\thispagestyle{plain}

\maketitle

\begin{abstract}
Semi-supervised machine learning models learn from
a (small) set of labeled training examples,
and a (large) set of unlabeled training examples.
State-of-the-art models can reach within a few percentage points of fully-supervised
training, while requiring 100$\times$ less labeled data.

We study a new class of vulnerabilities: poisoning attacks that modify the
\emph{unlabeled} dataset.
In order to be useful, unlabeled datasets are given strictly less 
review than labeled datasets, and adversaries can therefore poison them easily.
By inserting maliciously-crafted unlabeled examples
totaling just $0.1\%$ of the dataset size,
we can manipulate 
a model trained on this poisoned dataset to misclassify arbitrary
examples at test time (as any desired label).
Our attacks are highly effective across datasets 
and semi-supervised learning methods. 

We find that more accurate methods (thus more likely to be used)
are significantly more vulnerable
to poisoning attacks, and as such better training methods are
unlikely to prevent this attack.
To counter this we explore the space of defenses, and propose
two methods that mitigate our attack.
\end{abstract}

\section{Introduction}

One of the main limiting factors to applying machine learning in practice
is its reliance on large labeled datasets \cite{lecun2015deep}.
\emph{Semi-supervised learning} addresses this
by allowing a model to be trained on a
small set of (expensive-to-collect) labeled examples, and a large set of
(cheap-to-collect) unlabeled examples \cite{zhu2005semi,lee2013pseudo,oliver2018realistic}.
While semi-supervised machine learning has historically
been ``completely unusable'' \cite{vanhoucke2019quiet},
within the past two years these techniques have improved to the point
of exceeding the accuracy
of fully-supervised learning because of their ability to leverage additional data \cite{sohn2020fixmatch,xie2019unsupervised,xie2019selftraining}.
%

Because ``unlabeled data can often be obtained with minimal human labor'' \cite{sohn2020fixmatch}
and is often scraped from the Internet,
in this paper we perform an evaluation of
the impact of training on unlabeled data collected
from potential adversaries.
Specifically, we study \emph{poisoning attacks} where an adversary injects maliciously
selected examples in order to cause the learned model to 
mis-classify target examples.

Our analysis focuses on the key distinguishing factor
of semi-supervised learning: we exclusively poison
the \emph{unlabeled} dataset.
These attacks are especially powerful because the natural defense
that adds additional human review to the unlabeled data
eliminates the value of collecting unlabeled data (as opposed
to labeled data) in the first place.

We show that these unlabeled attacks are feasible by introducing
an attack that directly exploits the under-specification problem inherent
to semi-supervised learning.
State-of-the-art semi-supervised training works by first guessing labels
for each unlabeled example, and then trains on these guessed labels.
Because models must supervise their own training, we can inject a 
misleading sequence of examples into the unlabeled dataset that causes
the model to fool itself into labeling arbitrary test examples incorrectly.

We extensively evaluate our attack across multiple datasets and learning algorithms.
By manipulating just $0.1\%$ of the unlabeled examples, we can cause specific
targeted examples to become classified as any desired class.
In contrast, clean-label fully supervised poisoning attacks that achieve the same goal require poisoning $1\%$ of the labeled dataset.

Then, we turn to an evaluation of defenses to unlabeled dataset poisoning
attacks.
We find that existing poisoning defenses are a poor match for the problem
setup of unlabeled dataset poisoning.
To fill this defense gap, we propose two defenses that
partially mitigate our attacks by identifying and then removing
poisoned examples from the unlabeled dataset.

We make the following contributions:
\begin{itemize}
\item We introduce the first semi-supervised poisoning attack,
that requires control of just $0.1\%$ of the unlabeled data.
\item We show that there is a direct relationship between model
accuracy and susceptibility to poisoning: more accurate techniques
are significantly easier to attack.
\item We develop a defense to
perfectly separate the poisoned from clean examples by monitoring training dynamics.
\end{itemize}

\newpage

\section{Background \& Related Work}
%
%
%

\subsection{(Supervised) Machine Learning}

Let $f_\theta$ be a machine learning 
classifier (e.g., a deep neural network \cite{lecun2015deep})
parameterized by its \emph{weights} $\theta$.
While the \emph{architecture} of the classifier is human-specified,
the weights $\theta$ must first be \emph{trained} in order to solve the
desired task.

Most classifiers are trained through the process of Empirical Risk Minimization (ERM) \cite{vapnik1992principles}.
Because we can not minimize the \emph{true risk} (how well the classifier
performs on the final task), we construct a labeled
training set $\mathcal{X}$ to estimate the risk.
Each example in this dataset has an assigned label attached to it,
thus, let $(x,y) \in \mathcal{X}$ denote an input $x$ with the assigned
label $y$.
We write $c(x)=y$ to mean the true label of $x$ is $y$.
Supervised learning minimizes the aggregated loss
\[ \mathcal{L}(\mathcal{X}) = \sum_{(x,y) \in \mathcal{X}} L(f_\theta(x), y) \]
where we define the per-example loss $L$ as the task requires.
%
%
We denote training by the function $f_\theta \gets \mathcal{T}(f, \mathcal{X})$.

This loss function is non-convex; therefore, identifying the parameters $\theta$
that reach the global minimum is in general not possible.
However, the success of deep learning can be attributed to the fact that while
the global minimum is difficult to obtain,
we can reach high-quality local minima through performing 
stochastic gradient descent \cite{keskar2016large}.
%

%

\paragraph{Generalization.}
The core problem in supervised machine learning is ensuring that the learned
classifier generalizes to unseen data \cite{vapnik1992principles}.
A $1$-nearest neighbor classifier
achieves perfect accuracy on the training data, but likely will not
generalize well to \emph{test data},
another labeled dataset that is used to evaluate the
accuracy of the classifier.
Because most neural networks are heavily over-parameterzed\footnote{Models have enough parameters to memorize the training data \cite{zhang2016understanding}.},
a large area of research develops methods that 
to reduce the degree to which classifiers overfit to the training data \cite{srivastava2014dropout,keskar2016large}.

Among all known approaches, the best strategy today to increase generalization is
simply training on larger training datasets \cite{taori2020measuring}. 
Unfortunately, these large datasets are expensive to collect.
For example, it is estimated that ImageNet \cite{russakovsky2015imagenet} cost several million 
dollars to collect \cite{recht2019imagenet}.

To help reduce the dependence on labeled data,
\emph{augmentation} methods
artificially increase the size of a dataset by slightly perturbing
input examples.
For example, the simplest form of augmentation will 
with probability $0.5$ flip the image along the vertical
axis (left-to-right), and then shift the image vertically or
horizontally by a small amount.
State of the art augmentation methods \cite{devries2017improved,cubuk2019autoaugment,xie2019adversarial}
can help increase generalization slightly, but regardless of the augmentation
strategy, extra data is strictly more valuable to the extent that it is available \cite{taori2020measuring}.

\subsection{Semi-Supervised Learning}
\label{sec:ssl}

When it's the labeling process---and not the data collection process---that's expensive,
then Semi-Supervised Learning\footnote{We refrain from using the typical abbreviation, \emph{SSL}, in a security paper.}
can help alleviate the dependence of machine learning
on labeled data.
Semi-supervised learning changes the problem setup by introducing a new unlabeled dataset containing examples $u \in \mathcal{U}$.
The training process then becomes a new algorithm $f_\theta \gets \mathcal{T}_s(f, \mathcal{X}, \mathcal{U})$.
The unlabeled dataset typically consists of data drawn from a similar distribution
as the labeled data.
While semi-supervised learning has a long history \cite{scudder1965probability,mclachlan1975iterative,zhu2005semi,lee2013pseudo,oliver2018realistic}, recent techniques have made significant progress \cite{xie2019unsupervised,sohn2020fixmatch}.

Throughout this paper we study the problem of image classification, 
the primary domain where strong semi-supervised learning methods
exist \cite{oliver2018realistic}. \footnote{Recent work has explored alternate domains \cite{xie2019unsupervised,sohn2020simple,park2020improved}.}

\paragraph{Recent Techniques}
All state-of-the-art techniques from the past two years rely on the same setup \cite{sohn2020fixmatch}:
they turn the semi-supervised machine learning problem
(which is not well understood)
into a fully-supervised problem (which is very well understood).
To do this, these methods compute a ``guessed label''
$\hat{y} = f(u; \theta_i)$ for each unlabeled example $u \in \mathcal{U}$, and
then treat the tuple $(u, \hat{y})$ as if it were a labeled sample \cite{lee2013pseudo}, thus
constructing a new dataset $\mathcal{U}'$.
The problem is now fully-supervised, and we can perform training as if by computing
$\mathcal{T}(f, \mathcal{X} \cup \mathcal{U'})$.
Because $\theta_i$ is the model's current parameters,
note that we are
using the model's \emph{current} predictions to supervise its training for the
\emph{next} weights.

We evaluate the three current leading techniques: MixMatch \cite{berthelot2019mixmatch},
UDA \cite{xie2019unsupervised}, and
FixMatch \cite{sohn2020fixmatch}.
While they differ in their details on how they generate the guessed label,
and in the strategy they use to further regularize the model, all methods
generate guessed labels as described above.
These differences are not fundamental to the results of our paper, and 
we defer details to Appendix~A.

%
%

\paragraph{Alternate Techniques}
Older semi-supervised learning techniques are significantly less effective.
While FixMatch reaches $5\%$ error on CIFAR-10, none of these methods perform
better than a $45\%$ error rate---\textbf{nine times less accurate}.
%

Nevertheless, for completeness we consider older methods as well:
we include evaluations of Virtual Adversarial
Training \cite{miyato2018virtual}, PiModel \cite{laine2016temporal},
Mean Teacher \cite{tarvainen2017mean}, and Pseudo Labels \cite{lee2013pseudo}.
These older techniques often use a more ad hoc approach to learning, which
were later unified into a single solution.
For example, VAT \cite{miyato2018virtual} is built
around the idea of \emph{consistency regularization}: a model's predictions should
not change on perturbed versions of an input.
%
%
In contrast, Mean Teacher \cite{tarvainen2017mean} takes a different approach of
\emph{entropy minimization}: it uses prior 
models $f_{\theta_i}$ to train a later model $f_{\theta_j}$ (for $i<j$) and
find this additional regularization is helpful.
%

%
%

\subsection{Poisoning Attacks}

While we are the first to study poisoning
attacks on unlabeled data in semi-supervised learning,
there is an extensive line of work performing data poisoning attacks
in a variety of fully-supervised machine learning classifiers \cite{kearns1993learning,nelson2008exploiting,barreno2006can,koh2017understanding,turner2018clean,jagielski2018manipulating}
as well as un-supervised clustering attacks
\cite{biggio2013data,biggio2014poisoning,kloft2010online,kloft2012security}.

\paragraph{Poisoning labeled datasets.}
In a poisoning attack, an adversary either modifies existing examples
or inserts new examples into the training dataset in order to cause
some potential harm.
There are two typical attack objectives: indiscriminate and targeted poisoning.

In an \emph{indiscriminate} poisoning attack \cite{nelson2008exploiting,biggio2012poisoning}, the
adversary poisons the classifier to reduce its accuracy.
For example, Nelson \emph{et al.} \cite{nelson2008exploiting} 
modify $1\%$ of the training dataset
to reduce the accuracy of a spam classifier to chance.
%
%

\emph{Targeted} poisoning attacks \cite{nelson2008exploiting,chen2017targeted,koh2017understanding}, in comparison, aim to cause the specific (mis-)prediction 
of a particular example.
For deep learning models that are able to memorize the training dataset, 
simply mislabeling an example will cause a model to learn that incorrect label---however
such attacks are easy to detect.
As a result, \emph{clean label} \cite{shafahi2018poison} poisoning attacks inject only images that are correctly
labeled to the training dataset.
For instance, one state-of-the-art attack \cite{zhu2019transferable} modifies $1\%$ of the
training dataset in order to misclassify a CIFAR-10 \cite{krizhevsky2009learning}
test image.
Recent work \cite{liu2019unified} has studied
attacks that poison the \emph{labeled} dataset of a semi-supervised learning algorithm
to cause various effects.  This setting is simpler than ours, as
an adversary can control the labeling process.

Between targeted and indiscriminate attacks lies \emph{backdoor attack} \cite{gu2017badnets,turner2018clean,liu2020reflection}. Here, an adversary poisons a dataset so the model will
mislabel any image with a particular pattern applied,
but leaves all other images unchanged. We do not consider backdoor attacks in this paper.

%
%

\paragraph{Poisoning unsupervised clustering}
In \emph{unsupervised clustering},
there are \emph{no} labels, and the classifier's objective is to
group together similar classes without supervision.
Prior work has shown it is possible to poison clustering algorithms
by injecting unlabeled data to indiscriminately reduce model accuracy \cite{biggio2013data,biggio2014poisoning}.
This work constructs \emph{bridge} examples that connect independent clusters of examples.
By inserting a bridge connecting two existing clusters, the clustering
algorithm will group together both (original) clusters into one new cluster.
We show that a similar technique can be adapted to
\emph{targeted} misclassification attacks for semi-supervised learning.

Whereas this clustering-based work is able to analytically construct near-optimal
attacks \cite{kloft2010online,kloft2012security} for semi-supervised algorithms,
analyzing the dynamics of stochastic gradient descent is far more complicated.
Thus, instead of being able to derive an optimal strategy, we must perform
extensive experiments to understand how to form these bridges and understand
when they will successfully poison the classifier.

\subsection{Threat Model}

We consider a victim who trains a machine learning model on
a dataset with limited labeled examples (e.g., images, audio, malware, etc).
To obtain more \textbf{un}labeled examples, the victim scrapes (a portion of) the public
Internet for more examples of the desired type.
For example, a state-of-the-art Image classifier \cite{mahajan2018exploring}
was trained by scraping $1$ billion images off
of Instagram.
As a result, an adversary who can upload data to the Internet can control a portion
of the unlabeled dataset.

Formally, the unlabeled dataset poisoning adversary $\mathcal{A}$ constructs 
a set of poisoned examples
\[\mathcal{U}_p \gets \mathcal{A}(x^*, y^*, N, f, \mathcal{T}_s, \mathcal{X}').\]
The adversary receives the input $x^*$ to be poisoned,
the desired incorrect target label $y^* \ne c(x^*)$,
the number of examples $N$ that can be injected,
the type of neural network $f$,
the training algorithm $\mathcal{T}_s$,
and a subset of the labeled examples $\mathcal{X}' \subset \mathcal{X}$.

The adversary's goal is to poison the victim's model so that the model
$f_\theta \gets \mathcal{T}_s(\mathcal{X}, \mathcal{U} \cup \mathcal{U}_p)$
will classify the selected example as the desired target,
i.e., $f_\theta(x^*) = y^*$.
We require $\lvert \mathcal{U}_p\rvert < 0.01 \cdot \lvert \mathcal{U} \rvert$.
This value poisoning $1\%$ of the data has been consistently used in
data poisoning for over ten years \cite{nelson2008exploiting,biggio2012poisoning,shafahi2018poison,zhu2019transferable}.
(Interestingly, we find that in many settings we can succeed with just a $0.1\%$ poisoning ratio.)

To perform our experiments, we randomly select $x^*$ from among
the examples in the test set, and then sample a label $y^*$ randomly
among those that are different than the true label $c(x^*)$.
(Our attack will work for any desired example, not just an example in
the test set.)

\begin{figure*}

    \centering
    \hspace{.15em}
    \includegraphics[scale=.45]{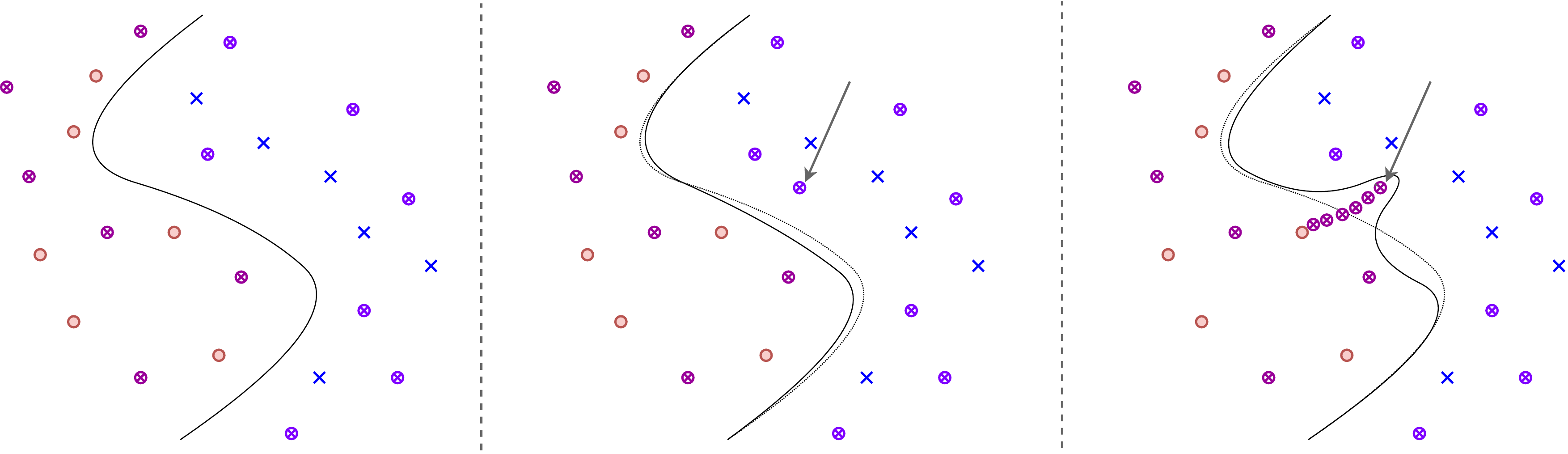}
         \captionsetup[sub]{size=normalsize}
     \begin{subfigure}{0.31\textwidth}
         \centering
         \caption{A classifier trained on a semi-supervised dataset
    of red $\odot$s, blue $\times$s, and \emph{unlabeled} $\otimes$s.
    During training the unlabeled $\otimes$s are given pseudo-labels such
    that the correct original decision boundary is learned.}
         \label{fig:five over x}
     \end{subfigure}
     \hfill
     \begin{subfigure}{0.31\textwidth}
         \centering
         \caption{ When inserting just one new \emph{unlabeled} poisoned example near the boundary,
    the model gives it the correct pseudo label
    of the blue $\times$s.
    The poisoning attempt fails, and the decision boundary remains largely unchanged.}
         \label{fig:five over x}
     \end{subfigure}
     \hfill
     \begin{subfigure}{0.31\textwidth}
         \centering
         \caption{By inserting a path of unlabeled examples,
    the classifier assigns every example in the path the
    pseudo-label of the nearby red $\odot$s.
    This moves the decision boundary to enclose the path,
    which makes these examples misclassified.}
         \label{fig:five over x}
     \end{subfigure}
     
  \caption{Decision boundary plot for semi-supervised learning during
  (a) normal training, (b) failed poisoning, and (c) our attack.}
    \label{fig:mainfig}
\end{figure*}

\section{Poisoning the Unlabeled Dataset}

We now introduce our semi-supervised poisoning attack,
which directly exploits the self-supervised \cite{scudder1965probability,mclachlan1975iterative}
nature of semi-supervised learning
that is fundamental to all state-of-the-art techniques.
Many machine learning vulnerabilities are attributed
to the fact that, 
instead of specifying \emph{how} a task should be
completed (e.g., look for three intersecting line segments),
machine learning specifies \emph{what} should be done 
(e.g., here are several examples of triangles)---and then we hope
that the model solves the problem in a reasonable manner.
However, machine learning models often
``cheat'', solving the task through
unintended means \cite{ilyas2019adversarial,geirhos2020shortcut}.

Our attacks show
this under-specification problem is exacerbated with semi-supervised learning:
now, on top of not specifying \emph{how} a task should be solved,
we do not even completely specify \emph{what} should be done.
When we provide the model with unlabeled examples
(e.g., here are a bunch of shapes),
we allow it to teach itself from this unlabeled
data---and hope it teaches
itself to solve the correct problem.

Because our attacks target the underlying principle behind semi-supervised
machine learning, they are general across techniques and are not
specific to any one particular algorithm.

\subsection{Interpolation Consistency Poisoning}

\paragraph{Approach.}
Our attack, Interpolation Consistency Poisoning,
exploits the above intuition.
Given the target image $x^*$, we begin by inserting it into the (unlabeled) portion
of the training data.
However, because we are unable to directly attach a label to this
example, we will cause the model itself to mislabel this example.
Following typical assumptions that the adversary has (at least partial)
knowledge of the training data \cite{shafahi2018poison,zhu2019transferable}%
\footnote{This is not a completely unrealistic assumption.
An adversary might obtain this knowledge through
a traditional information disclosure
vulnerability, a membership inference attack \cite{shokri2017membership},
or a training data extraction attack \cite{carlini2018secret}.}, we select any example in the
labeled dataset $x'$ 
that is correctly classified as the desired target label $y^*$, i.e., $c(x') = y^*$.
Then, we insert $N$ points 
\[\{x_{\alpha_i}\}_{i=0}^{N-1} = \text{interp}(x', x^*, \alpha_i)\]
where the $\text{interp}$ function is smooth along $\alpha_i \in [0,1]$ and
\begin{equation*}
    \text{interp}(x',x^*,0) = x' \,\,\,\,\,\,\,\,\,\,\,\,\,\,\,\,\,\,\,\,\,\,\,\,\,\,\,\, \text{interp}(x',x^*,1) = x^*.
\end{equation*}
This essentially connects the sample $x'$ to the sample $x^*$,
similar to the ``bridge'' examples from unsupervised clustering attacks \cite{biggio2014poisoning}.
Figure~\ref{fig:mainfig} illustrates the intuition behind this attack.

This attack relies on the reason that semi-supervised machine
learning is able to be so effective \cite{berthelot2019mixmatch,sohn2020fixmatch}.
Initially, only a small number of examples are labeled.
During the first few epochs of training, the model begins to classify these labeled
examples correctly: all semi-supervised training techniques include a standard
fully-supervised training loss on the labeled examples \cite{oliver2018realistic}.

As the confidence on the labeled examples grows, the neural network will also begin
to assign the correct label to any point nearby these examples.
There are two reasons that this happens:
First, it turns out that neural networks are
Lipschitz-continuous with a low constant (on average).
Thus, if $f(x) = y$ then ``usually'' we will have small $\epsilon$ perturbations
$f(x+\epsilon) = y+\delta$ for some small $\lVert\delta\rVert$.
\footnote{Note that this is true despite the existence of adversarial examples \cite{biggio2013evasion,szegedy2013intriguing},
which show that the \emph{worst-case} perturbations can change classification significantly.
Indeed, the fact that adversarial examples were considered ``surprising'' is exactly
due to this intuition.}
Second, because models apply data augmentation, 
they are already trained on perturbed inputs $x+\varepsilon$ generated by adding noise
to $x$; this reinforces the low average-case Lipschitz constant.

As a result, any nearby \emph{unlabeled} examples (that is, examples where $\lVert x_u - x\rVert$ is small)
will now also begin to be classified correctly with high confidence.
After the confidence assigned to these nearby unlabeled examples becomes 
sufficiently large, the training algorithms begins to treat these as if they
were labeled examples, too.
Depending on the training technique, the exact method by which the example become ``labeled''
changes.
UDA \cite{xie2019unsupervised}, for example, explicitly sets a confidence threshold at $0.95$ after which
an unlabeled example is treated as if it were labeled.
MixMatch \cite{berthelot2019mixmatch}, in contrast, performs a more complicated ``label-sharpening'' procedure that
has a similar effect---although the method is different.
The model then begins to train on this unlabeled example, and the process begins
to repeat itself.

When we poison the unlabeled dataset, this process happens in a much more controlled
manner.
Because there is now a path between the source example $x'$ and the target example $x^*$,
and because that
path begins at a labeled point, the model will assign the first unlabeled example $x_{\alpha_0}=x'$
the label $y^*$---its true and correct label.
As in the begnign setting,
the model will progressively assign higher confidence for this label on this example.
Then, the semi-supervised learning algorithms will
encourage nearby samples (in particular, $x_{\alpha_1}$)
to be assigned the same label $y^*$ as the label given to this first point $x_{\alpha_0}$.
This process then repeats.
The model assigns higher and higher likelihood to the example $x_{\alpha_1}$ to be classified as $y^*$,
which then encourages $x_{\alpha_2}$ to become classified as $y^*$ as well.
Eventually all injected examples $\{x_{\alpha_i}\}$ will be labeled the same way, as $y^*$.
This implies that finally, $f(x_{\alpha_0}) = f(x_{N-1}) = f(x^*) = y^*$ will be as well,
completing the poisoning attack.

\paragraph{Interpolation Strategy.}
It remains for us to instantiate the function $\text{interp}$.
To begin we use the simplest strategy: linear pixel-wise blending between the original example $x'$ and the
target example $x^*$.
That is, we define
\[\text{interp}(x',x^*,\alpha) = x' \cdot (1-\alpha) + x^* \cdot \alpha.\]
This satisfies the constraints defined earlier: it is smooth, and the boundary
conditions hold.
In Section~\ref{sec:gan} we will construct far more sophisticated interpolation strategies
(e.g., using a GAN \cite{goodfellow2014generative} to generate semantically meaningful interpolations);
for the remainder of this section we demonstrate the utility of a simpler strategy.

\paragraph{Density of poisoned samples.}
The final detail left to specify is how we choose the values of $\alpha_i$.
The boundary conditions $\alpha_0 = 0$ and $\alpha_{N-1}$ = 1 are
fixed, but how should we interpolate
between these two extremes?
The simplest strategy would be to sample completely linearly within the range $[0,1]$,
and set $\alpha_i = i/N$.
This choice, though, is completely arbitrary;
we now define what it would look like to provide different interpolation methods
that might allow for an attack to succeed more often.

Each method we consider works by first choosing a \emph{density functions} $\rho(x)$
that determines the sampling rate.
Given a density function, we first normalize it
\[\hat\rho(x) = \rho(x) \cdot \left(\int_0^1 \rho(x) \, dx\right)^{-1}\]
and then sample from it so that we sample $\alpha$ according to
\[\text{Pr}[p < \alpha < q] = \int_p^q \hat\rho(x)\,dx.\]

For example, the function $\rho(x) = 1$ corresponds to
a uniform sampling of $\alpha$ in the range $[0,1]$.
If we instead sample according to $\rho(x) = x$ then $\alpha$ will
be more heavily sampled near $1$ and less sampled near $0$,
causing more insertions near the target
example and fewer insertions near the original example.
Unless otherwise specified, in this paper we use the sampling function
$\rho(x) = 1.5-x$.
This is the function that we found to be most effective in an
experiment across eleven candidate density functions---see
Section~\ref{sec:eleven} for details.

\subsection{Evaluation}
\label{sec:evaluation}

We extensively validate the efficacy of our proposed attacks on
three datasets and seven semi-supervised learning algorithms.

\subsubsection{Experimental Setup}

\paragraph{Datsets}
We evaluate our attack on three datasets typically used in
semi-supervised learning:
\begin{itemize}
    \item CIFAR-10 \cite{krizhevsky2009learning} is the most studied
    semi-supervised learning dataset, with $50,000$ images
    from $10$  classes. 
    \item SVHN \cite{netzer2011reading} is a larger dataset of $604,388$ images
    of house numbers, allowing us to evaluate the efficacy
    of our attack on datasets with more examples.
    \item STL-10 \cite{coates2011analysis} is a dataset designed for
    semi-supervised learning. It contains just $1,000$ labeled images (at a higher-resolution, $96\times96$),
    with an additional $100,000$ unlabeled images drawn
    from a similar (but not identical) distribution, making it our most realistic dataset.
\end{itemize}

Because CIFAR-10 and SVHN were initially designed for fully-supervised
training,
semi-supervised learning research uses these dataset
by discarding the labels of all but a small
number of examples---typically just 40 or 250.

\paragraph{Semi-Supervised Learning Methods}
We 
perform most experiments on the three most accurate techniques:
MixMatch \cite{berthelot2019mixmatch}, UDA \cite{xie2019unsupervised}, and FixMatch \cite{sohn2020fixmatch},
with additional experiments on VAT \cite{miyato2018virtual}, Mean Teacher \cite{tarvainen2017mean}, Pseudo Labels \cite{lee2013pseudo}, and Pi Model \cite{laine2016temporal}.
The first three methods listed above are $3-10\times$
more accurate compared to the next four methods.
As a result we believe these will be most useful in the future,
and so focus on them.

We train all models with the same
$1.4$ million parameter ResNet-28 \cite{he2016deep} 
model that reaches $96.38\%$ fully-supervised accuracy.
This model has become the standard benchmark model for semi-supervised
learning \cite{oliver2018realistic} due to its relative simplicity and small size while also reaching
near state-of-the-art results \cite{sohn2020fixmatch}.
In all experiments we confirm that 
poisoning the unlabeled dataset maintains standard accuracy.

\paragraph{Experiment Setup Details.}
In each section below we answer several research questions;
for each \emph{experimental trial} we
perform 8 attack attempts and report the success rate.
In each of these 8 cases, we choose a new random source image
and target image uniformly at random.
Within each figure or table we re-use the same randomly selected images to
reduce the statistical noise of inter-table comparisons.

While it would clearly be preferable to run each experiment
with more than 8 trials,
semi-superivsed machine learning algorithms are extremely
slow to train: \emph{one} run of FixMatch takes $20$-GPU hours on CIFAR-10,
and over five days on STL-10.
Thus, with eight trials per experiment,
our evaluations represent hundreds of GPU-days of 
compute time.

\begin{figure}
         \centering
    \includegraphics[scale=.7]{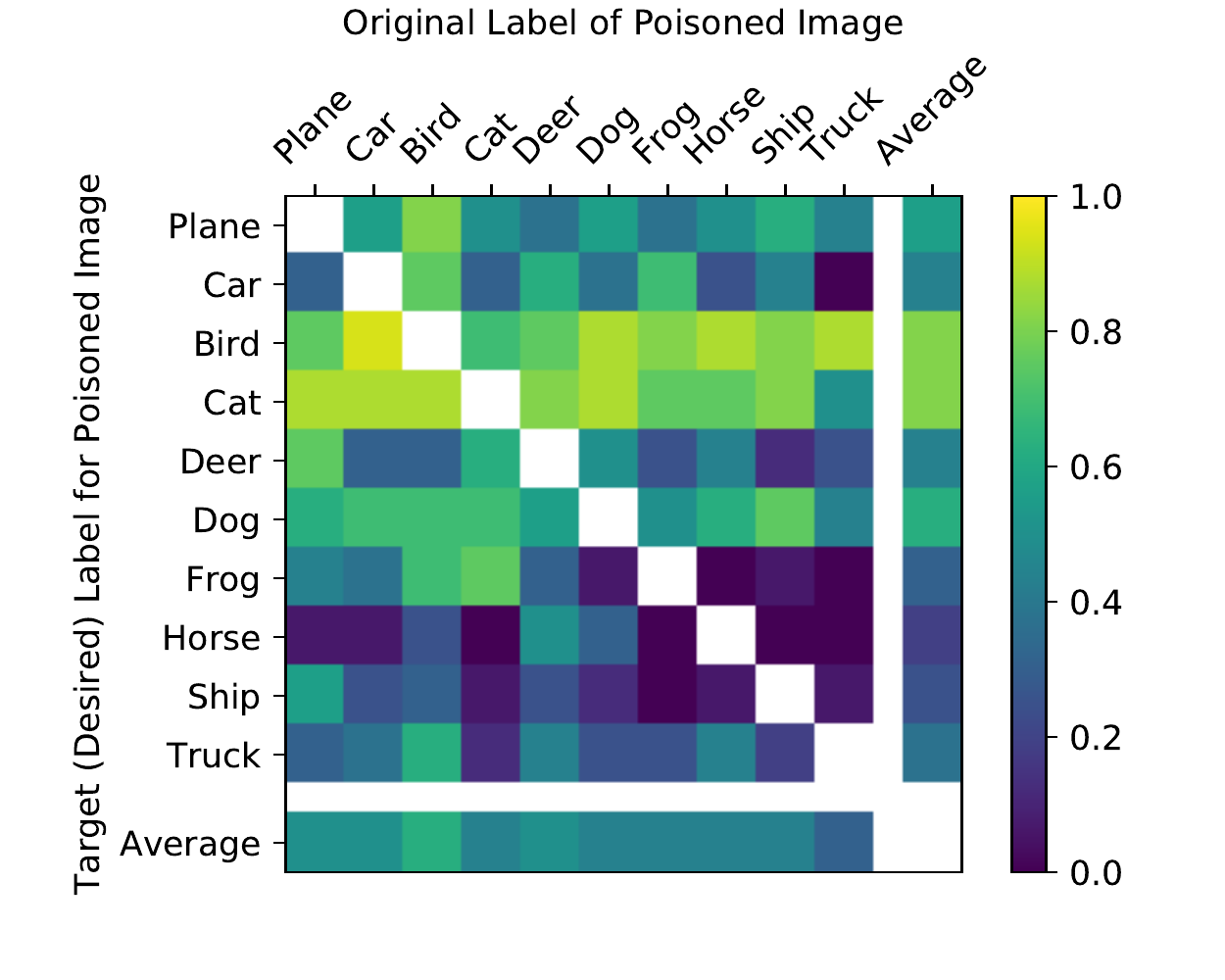}
         \caption{Poisoning attack success rate averaged across the ten CIFAR-10 classes.
         Each cell is the average of 16 trials.
         The original label of the (to-be-poisoned) image does not make attacks (much) easier or
  harder, but some target labels (e.g., horse) are harder to reach than others (e.g., bird).}
         \label{fig:grid1}
     \end{figure}

\subsubsection{Preliminary Evaluation}

We begin by demonstrating the efficacy of our attack on one
model (FixMatch) on one dataset (CIFAR-10) for one poison
ratio (0.1\%). 
Further sections will perform additional experiments that
expand on each of these dimensions.
When we run our attack eight different times with eight different
image-label pairs, we find that it
succeeds in seven of these cases.
However, as mentioend above, only performing eight trials is limiting---maybe
some images are easier or harder to successfully poison, or maybe some
images are better or worse source images to use for the attack.

\subsubsection{Evaluation across source- and target-image}

In order to ensure our attack remains consistently effective,
we now train an additional $40 \times 40$ models. 
For each model, we construct a different poison set by 
selecting $40$ source (respectively, target) images from the training (testing) sets,
with $4$ images from each of the $10$ classes.

We record each trial as a success if the target example
ends up classified as the desired label---or failure if not.
To reduce training time, we remove half of the unlabeled examples
(and maintain a $0.1\%$ poison ratio of the now-reduced-size dataset) and
train for a quarter the number of epochs.
Because we have reduced the total training, 
our attack success rate is
reduced to $51\%$ (future experiments will confirm
the baseline >80\% attack success rate found above).

Figure~\ref{fig:grid1} gives the attack success rate broken down by
the target image's true (original) label, and the desired poison label.
Some desired label such as ``bird'' or ``cat'' succeed in $85\%$ of cases,
compared to the most difficult label of ``horse'' that succeeds in $25\%$ of cases.

\begin{figure}
         \centering
        \includegraphics[scale=.7]{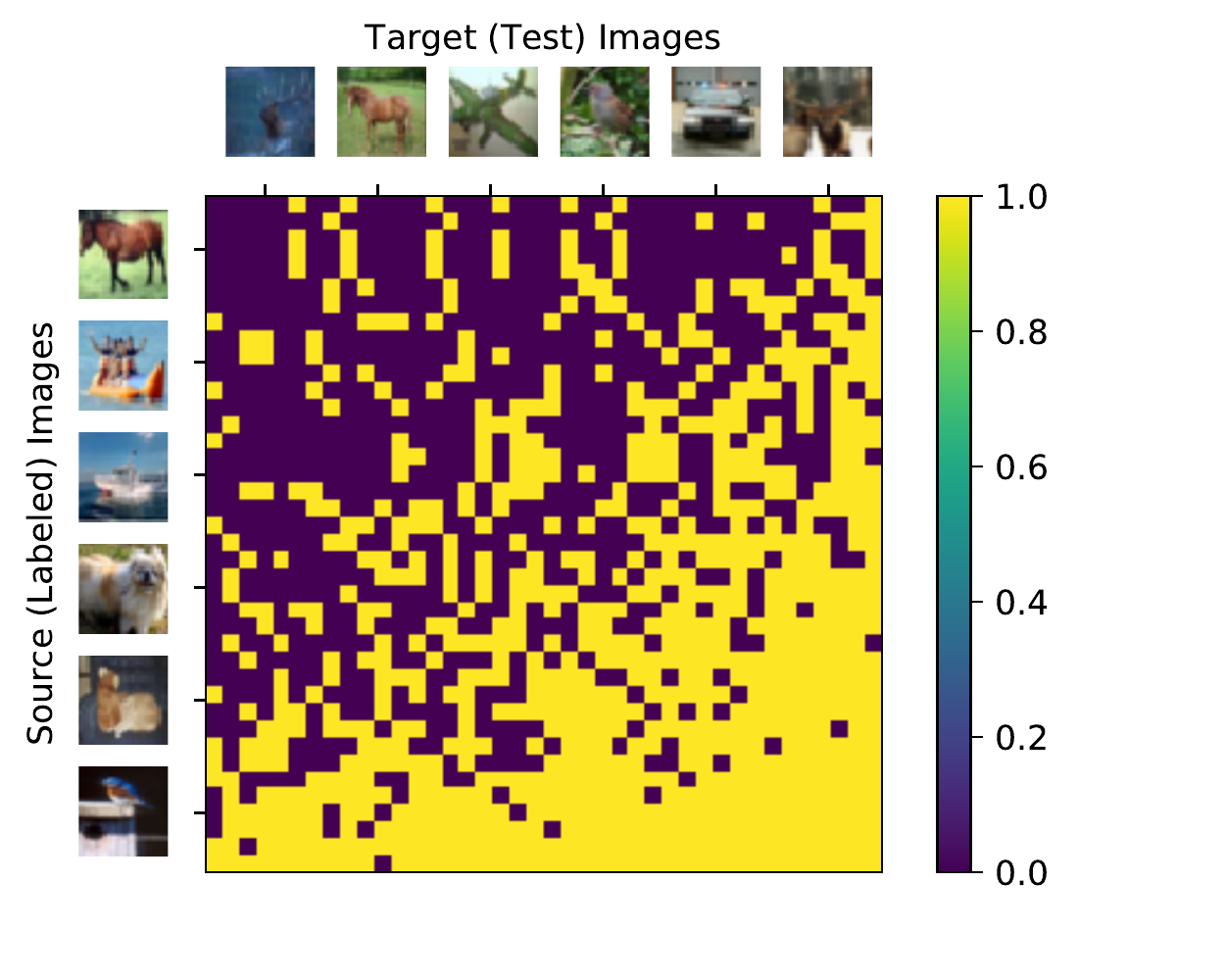}
         \caption{Poisoning attack success rate for all $40 \times 40$ source-target pairs;
         six (uniformly spaced) example images are shown on each axis.
         Each cell represents a single run of FixMatch poisoning that source-target pair,
         and its color indicates if the attack
         succeeded (yellow) or failed (purple).
         The rows and columns are sorted by average attack success rate.}
         \label{fig:grid2}
     \end{figure}

Perhaps more interesting than the aggregate statistics is considering 
the success rate on an image-by-image basis (see Figure~\ref{fig:grid2}). 
Some images (e.g., in the first column) can rarely
be successfully poisoned to reach the desired label, while other images
(e.g., in the last column) are easily poisoned.
Similarly, some source images (e.g., the last row) can poison almost any
image, but other images (e.g., the top row) are poor sources.
Despite several attempts, we have no explanation for why some
images are easier or harder to attack.

\begin{figure}
    \centering
    \includegraphics[scale=.8]{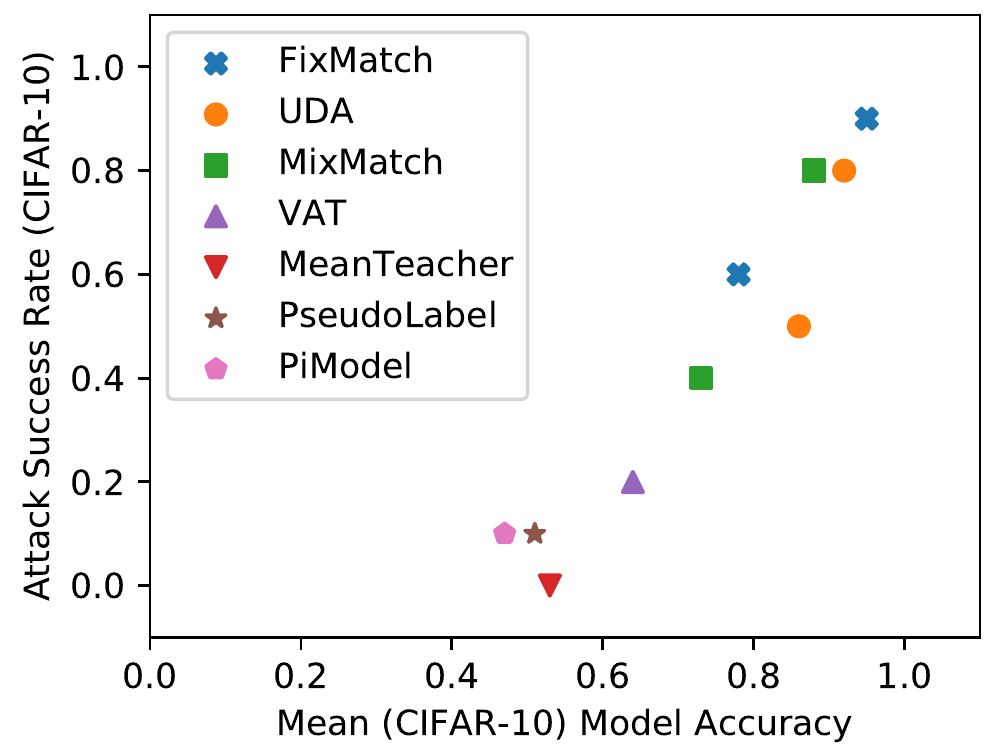}
    \caption{\textbf{More accurate techniques are more vulnerable.}
    Success rate of poisoning CIFAR-10 with 250 labeled examples and
    $0.2\%$ poisoning rate. Each point averages ten trained models.
    FixMatch, UDA, and MixMatch were trained under two evaluation settings, one standard
    (to obtain high accuracy) and one small-model to artificially reduce model accuracy.}
    \label{fig:controlacc}
\end{figure}

\begin{table*}
\centering
\begin{tabular}{l|rrr|rrr|rrr}
\toprule
     \textbf{Dataset} & \multicolumn{3}{c}{\textbf{CIFAR-10}}  & \multicolumn{3}{c}{\textbf{SVHN}}  & \multicolumn{3}{c}{\textbf{STL-10}} \\
     (\% poisoned) & 0.1\% & 0.2\% & 0.5\% & 0.1\% & 0.2\% & 0.5\% & 0.1\% & 0.2\% & 0.5\% \\
     \midrule
     MixMatch & 5/8 & 6/8 & 8/8    & 4/8 & 5/8 & 5/8 & 4/8 & 6/8 & 7/8 \\
     UDA      & 5/8 & 7/8 & 8/8    & 5/8 & 5/8 & 6/8 & - & - & - \\ 
     FixMatch & 7/8 & 8/8 & 8/8    & 7/8 & 7/8 & 8/8 & 6/8 & 8/8 & 8/8 \\ 
     \bottomrule
\end{tabular}
\vspace{.1em}
\caption{\textbf{Success rate of our poisoning attack across datasets and algorithms}, 
when poisoning between $0.1\%$ and $0.5\%$ of the unlabeled dataset.
CIFAR-10 and SVHN use 40 labeled examples, and STL-10 all 1000.
Our attack has a $67\%$ success rate when poisoning $0.1\%$ of the unlabeled dataset, and $91\%$ at $0.5\%$ of the unlabeled dataset (averaged across experiments).
}
\label{tab:resultmain}
\end{table*}

\subsubsection{Evaluation across training techniques}

The above attack shows that it is possible to poison
FixMatch on CIFAR-10 by poisoning $0.1\%$ of the training dataset.
We now broaden our argument by evaluating the attack success rate across seven
different training techniques--but again for just CIFAR-10.
As stated earlier, in all cases the poisoned models retain their original test accuracy compared to the benignly trained baseline on an unpoisoned dataset.

Figure~\ref{fig:controlacc} plots the main result of this experiment,
which compares the accuracy of the final trained model to 
the poisoning attack success rate.
The three most recent methods are all similarly vulnerable, with our attacks
succeeding over $80\%$ of the time.
When we train the four older techniques to the highest test accuracy they can
reach---roughly $60\%$---our poisoning attacks rarely succeed.

This leaves us with the question: why does our attack work less well on these
older methods?
Because it is not possible to artificially \emph{increase} the accuracy of worse-performing
techniques, we artificially \emph{decrease} the accuracy of the state-of-the-art techniques.
To do this, we train FixMatch, UDA, and MixMatch for fewer total steps of training
using a slightly smaller model in order to reduce their final accuracy to between $70-80\%$.
This allows us correlate the techniques accuracy
with its susceptibility to poisoning.

We find a clear relationship between the poisoning success rate
and the technique's accuracy.
We hypothesize this is caused by the better techniques
extracting more ``meaning'' from the unlabeled data.
(It is possible that, because we primarily experimented with recent techniques,
we have implicitly designed our attack to work better on these techniques. 
We believe the simplicity of our attack makes this unlikely.)
%
%
This has strong implications for the future. It suggests that developing
better training techniques is unlikely to prevent poisoning attacks---and will
instead make the problem worse.

\begin{table}
\centering
\begin{tabular}{l|rrr|rrr}
\toprule
     \textbf{Dataset} & \multicolumn{3}{c}{\textbf{CIFAR-10}}  & \multicolumn{3}{c}{\textbf{SVHN}} \\
     (\# labels) & 40 & 250 & 4000 & 40 & 250 & 4000 \\
     \midrule
     MixMatch & 5/8 & 4/8 & 1/8 & 6/8 & 4/8 & 5/8 \\ 
     UDA      & 5/8 & 5/8 & 2/8 & 5/8 & 4/8 & 4/8 \\ 
     FixMatch & 7/8 & 7/8 & 7/8 & 7/8 & 6/8 & 7/8 \\ 
     \bottomrule
\end{tabular}
\vspace{.1em}
\caption{Success rate of our attack when poisoning $0.1\%$ of the unlabeled dataset
when varying the number of labeled examples in the dataset.
Models provided with more labels are often (but not always) more robust to attack.}

\label{tab:labeleddata}
\end{table}

\subsubsection{Evaluation across datasets}
The above evaluation considers only one dataset at one poisoning ratio;
we now show that this attack is general across three datasets and three poisoning ratios.

Table~\ref{tab:resultmain} reports results for FixMatch, UDA, and MixMatch, as these
are the methods that achieve high accuracy. (We omit UDA on STL-10 because it was not
shown effective on this dataset in \cite{xie2019unsupervised}.)
Across all datasets, poisoning $0.1\%$ of the unlabeled data is sufficient
to successfully poison the model with probability at least $50\%$.
Increasing the poisoning ratio brings the attack success rate to near-$100\%$.

As before, we find that the
techniques that perform better are consistently more vulnerable across all experiment
setups.
For example, consider the poisoning success rate on SVHN.
Here again, FixMatch is more vulnerable to poisoning, with the attack
succeeding in aggregate for $20/24$ cases compared to $15/24$ for MixMatch.

%
%
%

%

\subsubsection{Evaluation across number of labeled examples}

Semi-supervised learning algorithms can be trained with a varying number of labeled
examples.
When more labels are provided, models typically perform more accurately.
%
We now investigate to what extent introducing more labeled examples impacts the
efficacy of our poisoning attack.
Table~\ref{tab:labeleddata} summarizes the results.
Notice that our prior observation comparing \emph{technique} accuracy to vulnerability
does not imply more accurate \emph{models}
are more vulnerable---with more training data, models are able to
learn with less guesswork and so become less vulnerable.
%

\begin{table}
\vspace{.5em}
    \centering
  \begin{tabular}{l|rrr}
    \toprule
      & \multicolumn{3}{c}{\textbf{CIFAR-10 \% Poisoned}} \\
    \textbf{Density Function}    & 0.1\% & 0.2\% & 0.5\% \\
    \midrule
    $(1-x)^2$        & 0/8 & 3/8 & 7/8 \\
    $\phi(x+.5)$ & 1/8 & 5/8 & 7/8 \\
    $\phi(x+.3)$ & 2/8 & 7/8 & 8/8 \\
    $x$        & 3/8 & 4/8 & 6/8 \\
    $x^4 + (1-x)^4$ & 3/8 & 5/8 & 8/8 \\
    $\sqrt{1-x}$   & 3/8 & 6/8 & 6/8 \\
    $x^2 + (1-x)^2$ & 4/8 & 5/8 & 8/8 \\
    $1$          & 4/8 & 6/8 & 8/8 \\
    $(1-x)^2+.5$     & 5/8 & 7/8 & 8/8 \\
    $1-x$          & 5/8 & 8/8 & 8/8 \\
    $1.5-x$       & 7/8 & 8/8 & 8/8 \\
    \bottomrule 
  \end{tabular}
   \vspace{.5em}
    
  \caption{Success rate of poisoning a semi-supervised machine learning model using different
  density functions to interpolate between the labeled example $x'$
  (when $\alpha=0$) and the target example $x^*$ (when $\alpha=1$).
  Higher values near $0$ indicate a more dense sampling near $x'$ and higher values
  near $1$ indicate a more dense sampling near $x^*$.
  Experiments conducted with FixMatch on CIFAR-10 using 40 labeled examples.
  }
  \label{tab:sample}
\end{table}

\subsubsection{Evaluation across density functions}
\label{sec:eleven}
All of the prior (and future) experiments in this paper use the same density
function $\rho(\cdot)$.
In this section we evaluate different choices to understand
how the choice of function impacts the attack success rate.

Table~\ref{tab:sample} presents these results. We evaluate each sampling method
across three different poisoning ratios.
As a general rule, the best sampling strategies sample slightly more
heavily from the source example, and less heavily from the target that will
be poisoned.
The methods that perform worst do not sample sufficiently densely near either
the source or target example, or do not sample sufficiently near
the middle.

For example, when we run our attack with the function
$\rho(x) = (1-x)^2$, then we sample frequently around the source image $x'$,
but infrequently around the target example $x^*$.
As a result, this density function fails at poisoning $x^*$ almost always,
because the density near $x^*$ is not high enough for the model's consistency
regularization to take hold.
We find that the label successfully propagates almost all the way to the final
instance (to approximately $\alpha=.9$) but the attack fails to change
the classification of the final target example.
Conversely, for a function like $\rho(x)=x$, the label propagation
usually fails near $\alpha=0$, but whenever it succeed at getting
past $\alpha>.25$ then it always succeeds at reaching $\alpha=1$.

Experimentally the best density function we identified was
$\rho(x) = 1.5-x$, which samples three times
more heavily around the source
example $x'$ than around the target $x^*$.

\subsection{Why does this attack work?}

Earlier in this section, and in Figure~\ref{fig:mainfig}, we provided
visual intuition why we believed our attack should succeed.
%
If our intuition is correct, we should expect two properties:

\begin{figure}
  \centering
  \,$\alpha=0.0$ \hspace{14em} $\alpha=1.0$
  \vspace{.2em} \\
  \includegraphics[scale=.685]{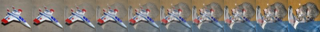}
  \includegraphics[scale=.75]{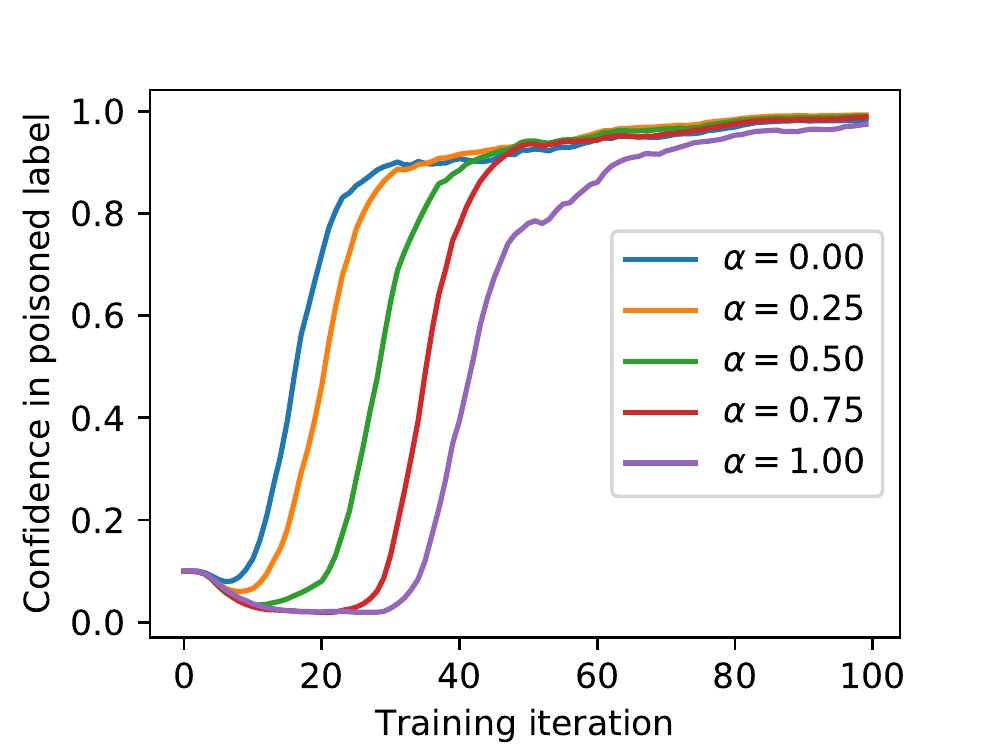}
  \caption{Label propagation of a poisoning attack over training epochs.
  The classifier begins by classifying the correctly-labeled
  source example $x'$ 
  (when $\alpha=0$; image shown in the upper left)
  as the poisoned label.
  This propagates to the interpolation $\alpha>0$ one by one,
  and
  eventually on to the final example $x^*$ 
  (when $\alpha=1$; image shown in the upper right).
  }
  \label{fig:curves}
\end{figure}

\begin{enumerate}
    \item As training progresses, the confidence of the model on each poisoned example
    should increase over time.
    \item The example $\alpha_0=0$ should become classified as the
    target label first, followed by $\alpha_1$, then $\alpha_2$, on to $\alpha_N=1$.
\end{enumerate}

We validate this intuition by
saving all intermediate model
predictions after every epoch of training.
This gives us a collection of model predictions for each epoch, for each 
poisoned example in the unlabeled dataset.

In Figure~\ref{fig:curves},
we visualize the predictions across a single FixMatch training run.%
\footnote{Shown above are the poisoned samples.
While blended images may look out-of-distribution,
Section~\ref{sec:gan} develops techniques to alleviate this.}
Initially, the model assigns all poisoned examples $10\%$ probability (because this is
a ten-class classification problem).
After just ten epochs, the model begins to classify the example $\alpha=0$ correctly.
This makes sense: $\alpha=0$ is an example in the labeled dataset and so it should
be learned quickly.
As $\alpha=0$ begins to be learned correctly, this prediction propagates to the samples
$\alpha>0$. In particular, the example $\alpha=.25$ (as shown) begins to become labeled
as the desired target.
This process continues until the poisoned example $x^*$ (where $\alpha=1$) begins to become
classified as the poisoned class label at epoch $80$.
By the end of training, all poisoned examples are classified as the desired target label.

\section{Extending the Poisoning Attack}

Interpolation Consistency Poisoning is effective at poisoning semi-supervised machine learning models
under this baseline setup,
but there are many opportunities for extending this attack.
Below we focus on three extensions that allow us to
\begin{enumerate}
\item \vspace{-.3em}attack without knowledge of any training datapoints,
\item \vspace{-.4em}attack with a more general interpolation strategy, and
\item \vspace{-.4em}attack transfer learning and fine-tuning.
\end{enumerate}

\subsection{Zero-Knowledge Attack}

Our first attack requires the adversary knows at least one example in the labeled dataset.
While there are settings where this is realistic \cite{shokri2017membership},
in general an adversary might have no knowledge of the labeled training dataset.
We now develop an attack that removes this assumption.

As an initial experiment, we investigate what would happen if we blindly
connected the target point $x^*$ with an arbitrary example $x'$ 
(not already contained within the labeled training dataset).
To do this we mount exactly our earlier attack without modification,
interpolating between an arbitrary unlabeled example $x'$ (that should belong to class $y^*$,
despite this label not being attached), and the target example $x^*$.

As we should expect,
across all trials, when we connect different source and target examples
the trained model consistently labels both
$x'$ and $x^*$ as the same final label.
Unexpectedly, we found that while the final label the model assigned was
rarely $y^*=c(x')$ (our attack objective label),
and instead
most often the final label was a label neither $y^*$ nor $c(x^*)$.

Why should connecting an example with label $c(x')$ and another example with label $c(x^*)$
result in a classifier that assigns neither of these two labels?
We found that the reason this happens is that, by chance, some intermediate image $x_\alpha$
will exceed the confidence threshold. Once this happens, both $x_{\alpha_1}$ and $x_{\alpha_{N-1}}$ become
classified as however $x_\alpha$ was classified---which often is not the label
as either endpoint.

In order to better regularize the attack process we provide additional support.
To do this, we choose additional images $\{\hat{x}_i\}$ and then connect each of
these examples to $x'$ with a path, blending as we do with the target example.

These additional interpolations make it more likely for $x'$ to be labeled
correctly as $y^*$, and when this happens, 
then more likely that the attack will succeed.

\paragraph{Evaluation}
Table~\ref{tab:unlab} reports the results of this poisoning attack.
Our attack success rate is lower, with roughly half of attacks succeeding
at $1\%$ of training data poisoned.
\emph{All} of these attacks succeed at making the target example $x^*$
becoming \emph{mislabeled} (i.e., $f(x^*) \ne c(x^*)$) even though it is not necessary
labeled as the desired target label.

\begin{figure}
\centering
\begin{tabular}{l|rr|rr}
\toprule
     \textbf{Dataset} & \multicolumn{2}{c}{\textbf{CIFAR-10}}  & \multicolumn{2}{c}{\textbf{SVHN}} \\
     (\% poisoned) & 0.5\% & 1.0\% & 0.5\% & 1.0\% \\
     \midrule
     MixMatch & 2/8 & 4/8 & 3/8 & 4/8 \\
     UDA      & 2/8 & 3/8 & 4/8 & 4/8 \\
     FixMatch & 3/8 & 4/8 & 3/8 & 5/8 \\
     \bottomrule
\end{tabular}
\vspace{.1em}
\caption{Success rate of our attack at poisoning the unlabeled dataset without knowledge
of any training examples. As in Table~\ref{tab:resultmain}, experiments are across three algorithms, but here across two datasets.}
\label{tab:unlab}
\end{figure}

\subsection{Generalized Interpolation}
\label{sec:gan}
When performing linear blending between the source and target example,
human visual inspection of the poisoned examples could identify them as out-of-distribution.
While it may be prohibitively expensive for a human to \emph{label} all of
the examples in the unlabeled dataset, it may not be expensive to
\emph{discard} examples that are clearly incorrect.
For example, while it may take a medical professional
to determine
whether a medical scan shows signs of some disease, any human
could reject images that were obviously not medical scans.

Fortunately
there is more than one way to interpolate between the source example $x'$ and the
target example $x^*$.
Earlier we used a linear pixel-space interpolation.
In principle, any interpolation strategy should remain effective---we now consider an alternate interpolation strategy as an example.

Making our poisoning attack inject samples that are not overly
suspicious therefore requires a more sophisticated interpolation
strategy.
Generative Adversarial Networks (GANs) \cite{goodfellow2014generative} are neural networks designed to 
generate synthetic images.
For brevity we omit details about training GANs as our results are independent
of the method used.
The \emph{generator} of a GAN is a function $g \colon \mathbb{R}^n \to \mathcal{X}$,
taking a \emph{latent} vector $z \in \mathbb{R}^n$ and returning an image.
GANs are widely used because of their ability to generate photo-realistic images.

One property of GANs is their ability to perform semantic
interpolation. 
Given two latent vectors $z_1$ and $z_2$ (for example, latent vectors corresponding
to a picture of 
person facing left and a person facing right), linearly interpolating between $z_1$
and $z_2$ will semantically interpolate
between the two images (in this case, by rotating the face from left to right).

For our attack, this means that it is possible to take our two images $x'$ and
$x^*$, compute the corresponding latent vectors $z'$ and $z^*$ so that
$G(z') = x'$ and $G(z^*) = x^*$, and then linearly interpolate to obtain
$x_i = G((1-\alpha_i) z' + \alpha_i z^*)$.
There is one small detail: in practice it is not always possible to obtain a
latent vector $z'$ so that exactly $G(z') = x'$ holds.
Instead, the best we can hope for is that $\lVert G(z') - x' \rVert$ is small.
Thus, we perform the attack as above interpolating between $G(z')$ and
$G(z^*)$ and then finally perform the small interpolation between
$x'$ and $G(z')$, and similarly $x^*$ and $G(z^*)$.

\paragraph{Evaluation.}
We use a DC-GAN \cite{radford2015unsupervised} pre-trained on CIFAR-10 to perform the interpolations.
We again perform the same attack as above, where we poison $1\%$ of the unlabeled examples by
interpolating in between the latent spaces of $z'$ and $z^*$.
Our attack succeeds in $9$ out of $10$ trials.
This slightly reduced attack success rate is a function of
the fact that while the two images are similarly far apart, the path taken is
less direct.

\subsection{Attacking Transfer Learning}
Often models are not trained from scratch but instead initialized from
the weights of a different model, and then \emph{fine tuned} on additional data \cite{pan2009survey}.
%
%
This is done either to speed up training via ``warm-starting'',
or when limited data is available (e.g., using ImageNet weights for cancer detection  \cite{esteva2017dermatologist}).

Fine-tuning is known to make attacks easier.
For example, if it's public knowledge that a model has been fine-tuned from a particular ImageNet
model, it becomes easier to generate adversarial examples on the fine-tuned model \cite{wang2018great}.
Similarly, adversaries might attempt to poison or backdoor a high-quality source model,
so that when a victim uses it for transfer learning their model becomes compromised as well \cite{yao2019latent}.

Consistent with prior work we find that it is easier
to poison models that are initialized from a pretrained model \cite{shafahi2018poison}.
The intuition here is simple.
The first step of our standard attacks waits for the model to assign
$x'$ the (correct) label $y^*$ before it can propagate to the target label $f(x^*) = y^*$.
If we know the initial model weights $\theta_{\text{init}}$,
then we can directly compute 
\[x' = \mathop{\text{arg min}}_{\delta\,\, :\,\, f_{\theta_{\text{init}}}(x^*+\delta)=y^*} \lVert \delta \rVert.\]
That is, we search for an example $x'$ that is nearby the desired target $x^*$ so that
the initial model
$f_{\theta_{\text{init}}}$ 
\emph{already}
assigns example $x'$ 
the label $y^*$.
Because this initial model assigns $x'$ the label $y^*$, then
the label will propagate to $x^*$---but because the two examples are closer, the
propagation happens more quickly.

\paragraph{Evaluation}
We find that this attack is even more effective than our baseline attack.
We initialize our model with a semi-supervised learning model trained on the
first $50\%$ of the unlabeled dataset to convergence.
Then, we provide this initial model weights $\theta_\text{init}$ to the adversary.
The adversary solves the minimization formulation above using standard gradient descent,
and then interpolates between that $x'$ and the same randomly selected $x^*$.
Finally, the adversary inserts poisoned examples into the second $50\%$ of the
unlabeled dataset, modifying just $0.1\%$ of the unlabeled dataset.

We resume training on this additional clean data (along with the poisoned
examples).
We find that, very quickly, the target example becomes successfully poisoned.
This matches our expectation: because the distance between the two examples is
very small, the model's decision boundary does not have to change by much in order
to cause the target example to become misclassified.
In $8$ of $10$ trials, this attack succeeds.

\subsection{Negative Results}

We attempted five different extensions of our attack that did not work.
Because we believe these may be illuminating, we now present each of these in turn.

\paragraph{Analytically computing the optimal density function}
In Table~\ref{tab:sample} we studied eleven different density functions.
Initially, we attempted to analytically compute the optimal density function,
however this did not end up succeeding.
Our first attempt repeatedly trained classifiers and performed binary search
to determine where along the bridge to insert new poisoned examples.
We also started with a dense interpolation of $500$ examples, and
removed examples while the attack succeeded.
Finally, we also directly computed the maximum distance $\epsilon$ so that training on
example $u$ would cause the confidence of $u+\delta$ (for $\lVert \delta \rVert_2 = \epsilon$)
to increase.

Unfortunately, each of these attempts failed for the same reason:
the presence or absence of one particular example is not independent of the other
examples.
Therefore, it is difficult to accurately measure the true influence of any particular
example, and greedy searches typically got stuck in local minima that required \emph{more}
insertions than just a constant insertion density with fewer starting examples.

\paragraph{Multiple intersection points}
Our attack chooses \emph{one} source $x'$ and
connects a path of unlabeled examples from that source $x'$ to the target $x^*$.
However, suppose instead that we selected multiple samples $\{x'_i\}_{i=1}^n$ and
then constructed paths from each $x'_i$ to $x^*$.
This might make it appear more likely to succeed: following the same intuition behind
our ``additional support'' attack, if one of the paths breaks,
one of the other paths might still succeed.

However, for the same insertion budget, experimentally we find it is
always better to use the entire budget on one single path from $x'$ to $x^*$ than to
spread it out among multiple paths.

\paragraph{Adding noise to the poisoned examples}
When interpolating between $x'$ and $x^*$ we experimented with adding point-wise
Gaussian or uniform noise to $x_\alpha$.
Because images are
discretized into $\{0, 1, \dots, 255\}^{hwc}$, it is possible that two sufficiently close
$\alpha, \alpha'$ will have $\text{discretize}(x_\alpha) = \text{discretize}(x_{\alpha'})$
even though $x_\alpha \ne x_{\alpha'}$.
By adding a small amount of noise, this property is no longer true, and therefore
the model will not see the same unlabeled example twice.

However, doing this did not improve the efficacy of the attack for small values of $\sigma$, and
made it worse for larger values of $\sigma$.
Inserting the same example into the unlabeled dataset two times was
more effective than just one time, because the model trains on that example twice
as often.

\paragraph{Increasing attack success rate.}
Occasionally, our attack gets ``stuck'', and $x'$ becomes classified as
$y^*$ but $x^*$ does not.
When this happens, the poisoned label only propagates \emph{part} of the way through
the bridge of poisoned examples.
That is, for some threshold $t$, we have that $x_{i < t} = y^*$ but
for $x_{i > t} \ne y^*$.
Even if $t=0.9$, and the propagation has made it almost all the way
to the final label, past a certain time in training the model's label assignments become
fixed and the predicted labels no longer change.
%
%
It would be interesting to better understand why these failures occur.

\paragraph{Joint labeled and unlabeled poisoning.}
Could our attack improve if
we gave the adversary the power to inject a single, correctly labeled,
poisoned example (as in a clean-label poisoning attack)?
We attempted various strategies to do this, ranging from inserting
out-of-distribution data \cite{russakovsky2015imagenet}
to mounting a Poison Frog attack \cite{shafahi2018poison}.
However, none of these ideas worked better than just choosing
a good source example as determined in Figure~\ref{fig:grid2}.
Unfortunately, we currently do not have a technique to predict which
samples will be good or bad sources (other than brute force training).

\section{Defenses}

We now shift our focus to preventing the poisoning attack we have just
developed.
While we believe existing defense are not well suited to defend against
our attacks, we believe that by combining automatic techniques to identify potentially-malicious
examples, and then manually reviewing these limited number of cases, 
it may be possible to prevent this attack.

\subsection{General-Purpose Poisoning Defenses}
While there are a large class of defenses to indiscriminate
poisoning attacks \cite{jagielski2018manipulating,steinhardt2017certified,kearns1993learning,candes2011robust,chen2013robust}, there are many fewer defenses that prevent the targeted poisoning attacks we study.
We briefly consider two defenses here.

Fine-tuning based defenses \cite{liu2018fine} take a (potentially poisoned) model
and fine-tune it on clean, un-poisoned data.
These defenses hope (but can not guarantee) that doing this will remove any
unwanted memorization of the model.
In principle these defenses should work as well on our setting as any other if
there is sufficient data available---however, because semi-supervised
learning was used in the first case, it is unlikely there will exist a large,
clean, un-poisoned dataset.

Alternatively, other defenses \cite{hong2020effectiveness} alter the training process
to apply differentially private SGD \cite{abadi2016deep} 
in order to mitigate the ability of the model to memorize training examples.
However, because the vulnerability of this defense scales exponentially 
with the number of poisoned examples,
these defenses are only effective at preventing extremely limited poisoning attacks 
that insert fewer than three or five examples.

Our task and threat model are sufficiently different from these prior defenses
that they are a poor fit for our problem domain: the threat models do not closely align.

\subsection{Dataset Inspection \& Cleaning}
We now consider two defenses tailored specifically to
prevent our attacks.
While it is undesirable to pay a human to manually inspect the entire
unlabeled dataset (if this was acceptable then the entire dataset might as well
be labeled),
this does not preclude any human review.
Our methods directly process the
unlabeled dataset and filter out a small subset of the examples to have reviewed by
a human for general correctness
(e.g., ``does this resemble anything like a dog?'' compared
to ``which of the 100+ breeds of dog in the ImageNet dataset is this?'').

\paragraph{Detecting pixel-space interpolations}
Our linear image blending attack is trivially detectable.
Recall that for this attack we set $x_{\alpha_i} = (1-\alpha_i) \cdot x' + \alpha_i \cdot x^*$.
Given the unlabeled dataset $\mathcal{U}$, this means that there will exist at least three
examples $u,v,w \in \mathcal{U}$ that are colinear in pixel space.
For our dataset sizes, a trivial trial-and-error sampling identifies the
poisoned examples in under ten minutes on a GPU.
%
%
While effective for this particular attack,
it can not, for example, detect our GAN latent space attack.

We can improve on this to detect arbitrary interpolations.
\emph{Agglomerative Clustering} \cite{ward1963hierarchical} creates clusters of similar
examples (under a pixel-space $\ell_2$ norm, for example).
Initially every example is placed into its own set.
Then, after computing the pairwise distance between all sets,
the two sets with minimal distance are merged to form one set.
The process repeats
until the smallest distance exceeds some threshold.

Because our poisoned examples are all similar in pixel-space to each other, it is
likely that they will be all placed in the same cluster.
Indeed, running a standard implementation of agglomerative clustering \cite{pedregosa2011scikit} 
is effective at identifying the poisoned examples in our attacks.
Thus, by removing the largest cluster, we can completely prevent this attack.

The inherent limitation of this defenses is that
it assume that the defender can create a useful distance function.
Using $\ell_2$ distance above worked because our attack performed
pixel-space blending.
However, if the adversary inserted examples that applied color-jitter, or
small translations, this defense would no longer able to identify the cluster
of poisoned examples.
This is a cat-and-mouse game we want to avoid.

\subsection{Monitoring Training Dynamics}
\label{sec:defense}

Unlike the prior defenses that inspect the dataset directly to detect
if an example is poisoned or not, we now develop a second
strategy that predicts if an example is poisoned by how it
impacts the training process.

Recall the reason our attack succeeds: semi-supervised learning
slowly grows the correct decision boundary out from the initial
labeled examples towards the ``nearest neighbors'' in the unlabeled examples,
and continuing outwards.
The guessed label of each unlabeled example will therefore be influenced by
(several) other unlabeled examples.
For benign examples in the unlabeled set, we should expect that they will be
influenced by many other unlabeled examples simultaneously, of roughly
equal importance.
%
%
However, by construction, our poisoned examples are designed to predominantly
impact the prediction of the other poisoned examples---and not be affected by,
or affect, the other unlabeled examples.

We now construct a defense that relies on this observation.
By monitoring the training dynamics of
the semi-supervised learning algorithm, we isolate out (and then remove)
those examples that are influenced by only a few other examples.

\paragraph{Computing pairwise influence}
What does it mean for one example to influence the training of another example?
In the context of fully-supervised training, there are rigorous definitions
of \emph{influence functions} \cite{koh2017understanding} 
that allow for one to (somewhat) efficiently compute
the training examples that maximally impacted a given prediction.
However, our influence question has an important difference: we ask not what
training points influenced a given test point, but what
(unlabeled) training points influenced another (unlabeled) training point.
We further find that it is not
necessary to resort to such sophisticated approaches, and a simpler (and $10-100\times$ faster)
method is able to effectively recover influence.

While we can not completely isolate out correlation and causation
without modifying the training process, we make 
the following observation that is fundamental to this defense:
\begingroup
\addtolength\leftmargini{0.2in}
\begin{quote}
    If example $u$ influences example $v$, then when the prediction of $u$
    changes, the prediction of $v$ should change in a similar way.
\end{quote}
\endgroup
After every epoch of training, we record the model's predictions on each
unlabeled example.
Let $f_{\theta_i}(u_j)$ denote the model's prediction vector after
epoch $i$ on the $j$th unlabeled example.
For example, in a binary decision task, if at epoch $6$ the model assigns
example $u_5$ class $0$ with probability $.7$ and class $1$ with $.3$,
then $f_{\theta_6}(u_5) = \begin{bmatrix} .7 & .3 \end{bmatrix}$.
From here, define 
\[\partial f_{\theta_i}(u_j) = f_{\theta_{i+1}}(u_j) - f_{\theta_{i}}(u_j)\]
with subtraction taken component-wise.
That is, $\partial f$ represents the difference in the models predictions
on a particular example from one epoch to the next.
This allows us to formally capture the intuition for ``the prediction changing''.

Then, for each example, we let
\[\mu_j^{(a,b)} = \begin{bmatrix} \partial f_{\theta_a}(u_j) & \partial f_{\theta_{a+1}}(u_j) & \dots & \partial f_{\theta_{b-1}}(u_j) &  f_{\theta_{b}}(u_j) \end{bmatrix} \]
be the model's collection of prediction deltas from epoch $a$ to epoch $b$.
We compute the influence of example $u_i$ on $u_j$ as
\[\text{Influence}(u_i, u_j) = \lVert \mu_i^{(0,K-2)} - \mu_j^{(1,K-1)} \rVert_2^2.\]
%

That is, example $u_i$ influences example $u_j$ if when example $u_i$ makes
a particular change at epoch $k$, then example $u_j$ makes a similar change
in the next epoch---because it has been influenced by $u_i$.
This definition of influence is clearly a simplistic approximation, and is
unable to capture sophisticated relationships between examples.
%
%
We nevertheless find that this definition of influence is useful.
%
%

\begin{figure}
\centering
  \includegraphics[scale=.8]{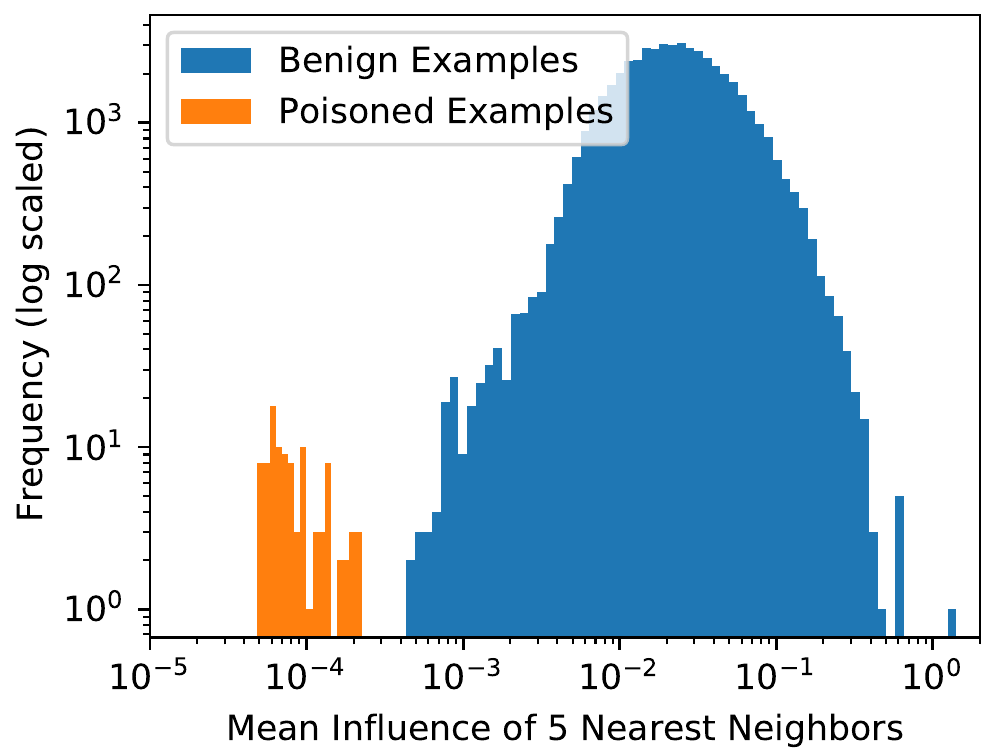}
  \caption{Our training-dynamics defense perfectly separates the inserted poisoned examples from the benign unlabeled
  examples on CIFAR-10 for a FixMatch poisoned model.
  Plotted is the frequency of the influence value across the unlabeled examples.
  Benign unlabeled examples are not heavily influenced by their nearest neighbors (indicated
  by the high values), but
  poisoned examples are highly dependent on the other poisoned examples (indicated by
  the low values).}
  \label{fig:defense}
\end{figure}
\paragraph{Identifying poisoned examples}
For each example in the unlabeled training set, we compute the average influence of
the $5$ nearest neighbors
\[\text{avg influence}(u) = {1 \over 5}\sum_{v \in \mathcal{U}} \text{Influence}(u,v) \cdot \mathds{1}[\text{close}_5(u,v)] \]
where $\text{close}_5(u,v)$ is true if $v$ is one of the $5$ closest (by influence) neighbors to $u$.
(Our result is not sensitive to the arbitrary choice of $5$; values from $2$ to $10$ perform similarly.)

\paragraph{Results.}
This technique perfectly separates the clean and poisoned examples for each task
we consider.
In Figure~\ref{fig:defense} we plot a representative histogram of influence values
for benign and poisoned examples; here we train a FixMatch model poisoning $0.2\%$
of the CIFAR-10 dataset and $40$ labeled examples.
The natural data is well-clustered with an average influence of $2 \cdot 10^{-2}$,
and the injected poisoned examples \emph{all} have an influence lower than
$2 \cdot 10^{-4}$, with a mean of $3 \cdot 10^{-5}$.
Appendix~B shows 8 more plots for additional runs of FixMatch and MixMatch on CIFAR-10,
and SVHN.

When the attack itself fails to poison the target class, it is still possible to identify
all of the poisoned examples that have been inserted (i.e., with a true positive rate of $100\%$),
but the false positive rate increases slightly to $0.1\%$.
In practice, all this means the defender should collect a few percent
more unlabeled examples
more than are required so that any malicious examples can be removed.
Even if extra training data is not collected, training on $99.9\%$ of the unlabeled dataset
with the false positives removed does not statistically significantly reduce clean model accuracy.

Thus, at cost of doubling the training time---training once with poisoned examples,
and a second time after removing them---it is possible to completely prevent our attack.
Multiple rounds of this procedure might improve its success rate further if not
all examples can be removed in one iteration.

\begin{figure}
    \centering
    \includegraphics[scale=.8]{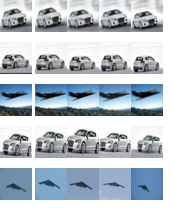}
    \caption{\textbf{Our defenses's near false positives are duplicated images.}
    The left-most column contains five images from the CIFAR-10 
    unlabeled dataset that our defense identifies as \emph{near false positives}.
    At right are the four next-most-similar images from the CIFAR-10 unlabeled set
    as computed by our average influence definition.
    All of these similar images are visual (near-)duplicates of the first.}
    \label{fig:fp}
\end{figure}

\paragraph{Examining (near) false positives.}
Even the near false positives
of our defense are insightful to analyze.
In Figure~\ref{fig:fp} we show five (representative) images of the benign
examples in the CIFAR-10 training dataset that our defense almost rejects as
if they were poisoned examples.

Because these examples are all nearly identical, they heavily influence each
other according to our definition.
When one examples prediction changes, the other examples predictions are likely
to change as well.
This also explains why removing near-false-positives does not reduce model accuracy.

\paragraph{Counter-attacks to these defenses.} No defense is full-proof, and
this defense is no different.
It is likely that future attacks will defeat this defense, too.
However, we believe that defenses of this style are a promising direction that 
(a) serve as a strong baseline for defending against unlabeled
dataset poisoning attacks, and (b) could be extended in future work.

\section{Conclusion}

Within the past years, semi-supervised learning has gone
from ``completely unusable'' \cite{vanhoucke2019quiet} to nearly as accurate
as the fully-supervised baselines despite using $100\times$ less labeled data. 
%
%
However, using semi supervised learning in practice will require
understanding what new vulnerabilities will arise as a result of
training on this under-specified problem.

In this paper we study the ability for an adversary to poison semi-supervised
learning algorithms by injecting unlabeled data.
As a result of our attacks, production systems will not be able to just take all 
available unlabeled data, feed it into a classifier, and hope for the best.
If this is done,
an adversary will be able to cause specific, targeted misclassifications.
Training semi-supervised learning algorithms on real-world data 
will require defenses tailored to preventing poisoning attacks whenever
collecting data from untrusted sources.

Surprisingly, we find that \emph{more accurate} semi-supervised learning methods
are \emph{more vulnerable} to poisoning attacks.
Our attack never succeeds on MeanTeacher because it has a $50\%$ error rate on
CIFAR-10; in contrast, FixMatch reaches a $5\%$ error rate
and as a result is easily poisoned.
This suggests that simply waiting for more accurate methods not only won't solve
the problem, but may even make the problem worse as future methods become more accurate.
%
%
%

Defending against poisoning attacks also can not be
achieved through extensive use of human review---doing so would reduce or
eliminate the only reason to apply semi-supervised learning in the first place.
Instead, we study defenses that isolate a small fraction of examples
that should be reviewed or removed.
Our defenses are effective at preventing the poisoning attack we present,
and we believe it will provide a strong baseline by which future work can be evaluated.

%
More broadly, we believe that or analysis highlights that
the recent trend where machine learning systems are trained
on any and all available data, without
regard to its quality or origin,
might introduce new vulnerabilities.
Similar trends have recently been observed in other domains;
for example, neural language models trained on unlabeled
data scraped from the Internet can be
poisoned to perform targeted mispredictions \cite{schuster2021you}.
We expect that other forms of unlabeled training, such as 
\emph{self}-supervised learning, will be similarly vulnerable to these
types of attacks.
We hope our analysis will allow future work to perform additional study
of this phenomenon in other settings where uncurated data is
used to train machine learning models.

\section*{Acknowledgements}
We are grateful to Andreas Terzis, David Berthelot, and the anonymous
reviewers for the discussion, suggestions, and feedback that significantly improved this paper.

{\footnotesize
\bibliographystyle{IEEEtranS}
\bibliography{IEEEabrv,paper}
}

\appendix

\section{Semi-Supervised Learning Methods Details}

We begin with a description of three state-of-the-art methods:
\begin{itemize}
    \item \textbf{MixMatch} \cite{berthelot2019mixmatch} generates a label guess for each unlabeled image,
    and \emph{sharpens} this distribution by increasing the softmax temperature.
    During training MixMatch penalizes the $L_2$ loss between this sharpened 
    distribution and another prediction on the unlabeled example.
    As additional regularization, MixMatch applies MixUp \cite{zhang2017mixup},
    weight decay \cite{krogh1992simple}, an exponential moving average of model parameters,
    trained with the Adam optimizer \cite{kingma2014adam}.
    
    \item \textbf{UDA} \cite{xie2019unsupervised} at its core behaves similarly to MixMatch, 
    and generates label guesses and matches the sharpened
    guess as MixMatch does.
    Instead of applying standard weak augmentation, UDA
    was the first method to show that strong augmentation
    can effectively increase the accuracy of semi-supervised learning algorithms.
    UDA again also contains a number of implementation details, including
    a cosine decayed learning rate, an additional KL-divergence loss regularizer,
    and training signal annealing.
    
    \item \textbf{FixMatch} \cite{sohn2020fixmatch} is a simplification of MixMatch and UDA.
    FixMatch again generates a guessed label and trains on this label, however
    it uses hard pseudo-labels instead of a sharpened label distribution and
    only trains on examples that exceed a confidence threshold.
    By carefully tuning parameters, FixMatch is able to remove from MixMatch the
    MixUp regularization and Adam training, and remove from UDA all of the
    details above (KLD loss, training signal annealing).
    Because it is a simpler methods, it is easier to determine optimal
    hyperparameter choices and it is able to achieve higher accuracy.
\end{itemize}

We also describe the four prior methods:
\begin{itemize}
    \item \textbf{Pseudo Labels} \cite{lee2013pseudo} is one of the early semi-supervised learning
    methods that gave high accuracy, and is built on by most others.
    This technique introduced the label guessing procedure, and exclusively
    works by generating a guessed label for each unlabeled example, and then training
    on that guessed label.
    
    \item \textbf{Virtual Adversarial Training} \cite{miyato2018virtual} proceeds by guessing a label
    for each unlabeled example using the current model weights.
    Then, it minimizes two terms.
    First, it minimizes the entropy of the guessed label, to encourage confident
    predictions on unlabeled examples.
    Then, it generates an adversarial perturbation $\delta$ for each unlabeled
    example and encourages the prediction $f(x_u)$ to be similar to the
    prediction $f(x_u + \delta)$.
    
    \item \textbf{Mean Teacher} \cite{tarvainen2017mean} again generates a pseudo label for each unlabeled
    example and trains on this pseudo label.
    However, instead of generating a pseudo label using the current model weights,
    it generates the pseudo label using an exponential moving average of prior
    model weights.
    
    \item \textbf{PiModel} \cite{laine2016temporal} relies heavily on \emph{consistency regularization}.
    This technique introduces the augmentation-consistency as used in prior
    techniques, however this paper uses network dropout instead of input-space
    perturbations to regularize the model predictions.
\end{itemize}

\newpage
\section{Additional Defense Figures}

\begin{figure}[H]
\centering
  \includegraphics[scale=.4]{figs/fixmatch0.pdf}
  \includegraphics[scale=.4]{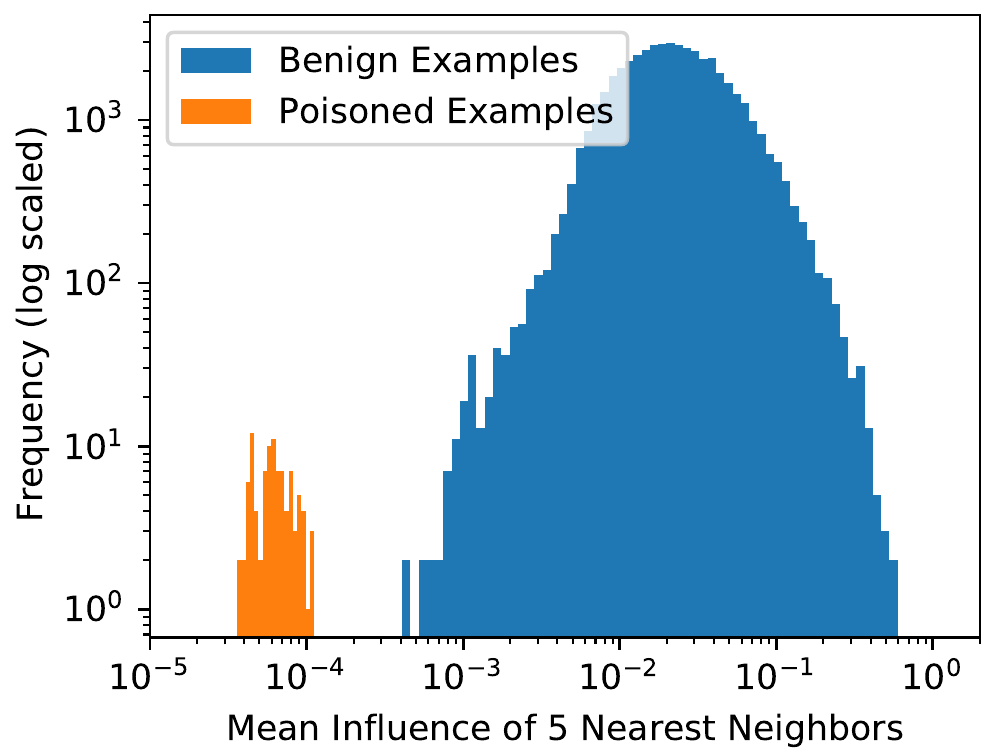}
  \caption{Our defense reliably detects poisoning attacks using FixMatch on CIFAR-10.
  In all cases, we perfectly separate the standard training data from the injected
  poisoned examples.}
  \label{fig:defensex0}
\end{figure}

\begin{figure}[H]
\centering
  \includegraphics[scale=.4]{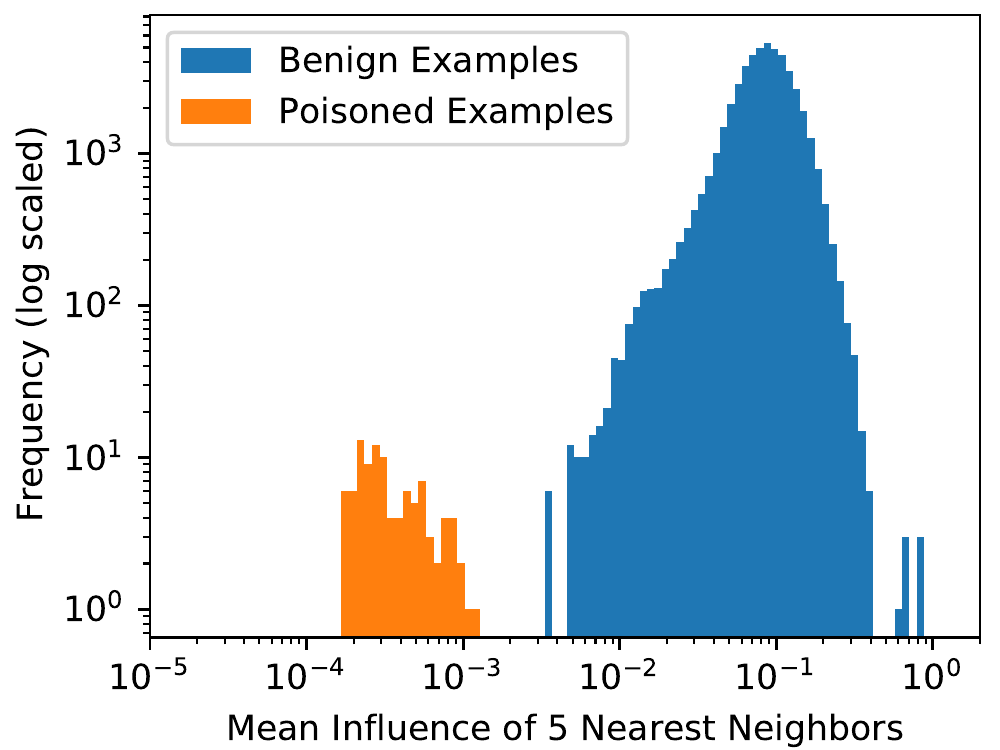}
  \includegraphics[scale=.4]{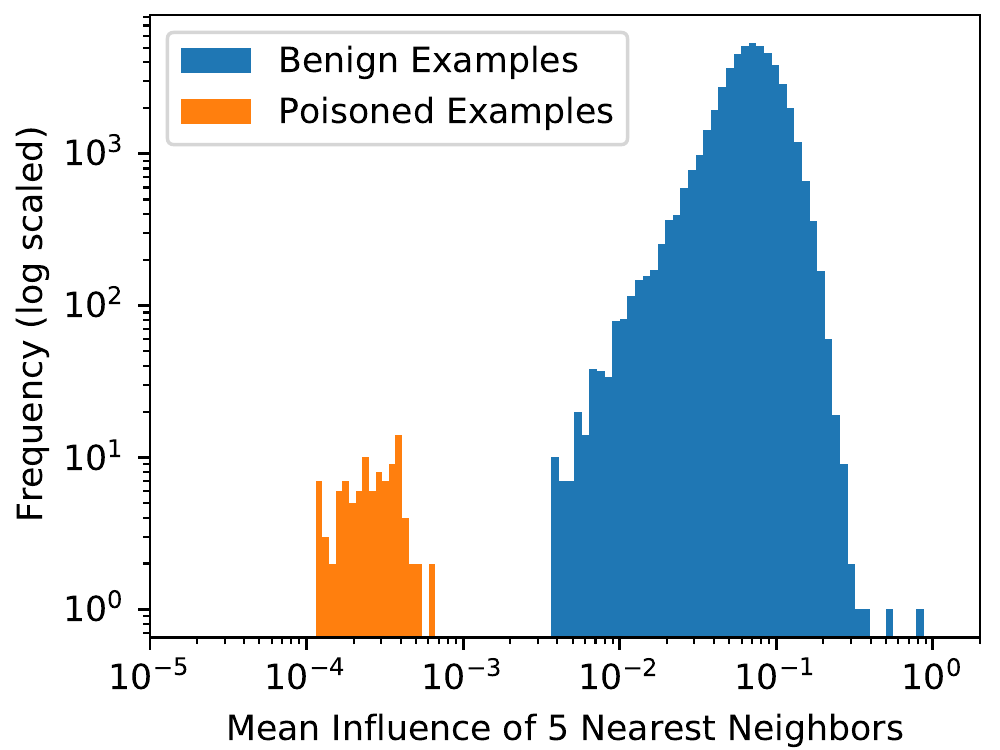}
  \caption{Our defense reliably detects poisoning attacks using MixMatch on CIFAR-10.
  In all cases, we perfectly separate the standard training data from the injected}
  \label{mixmatchcifar10}
\end{figure}

\begin{figure}[H]
\centering
  \includegraphics[scale=.4]{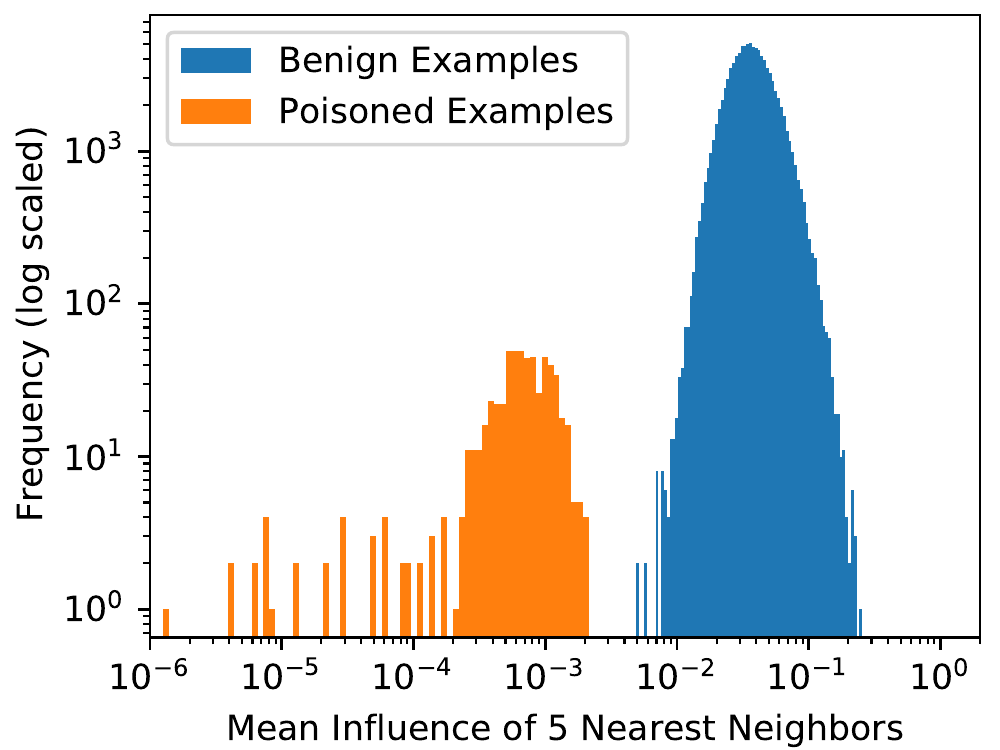}
  \includegraphics[scale=.4]{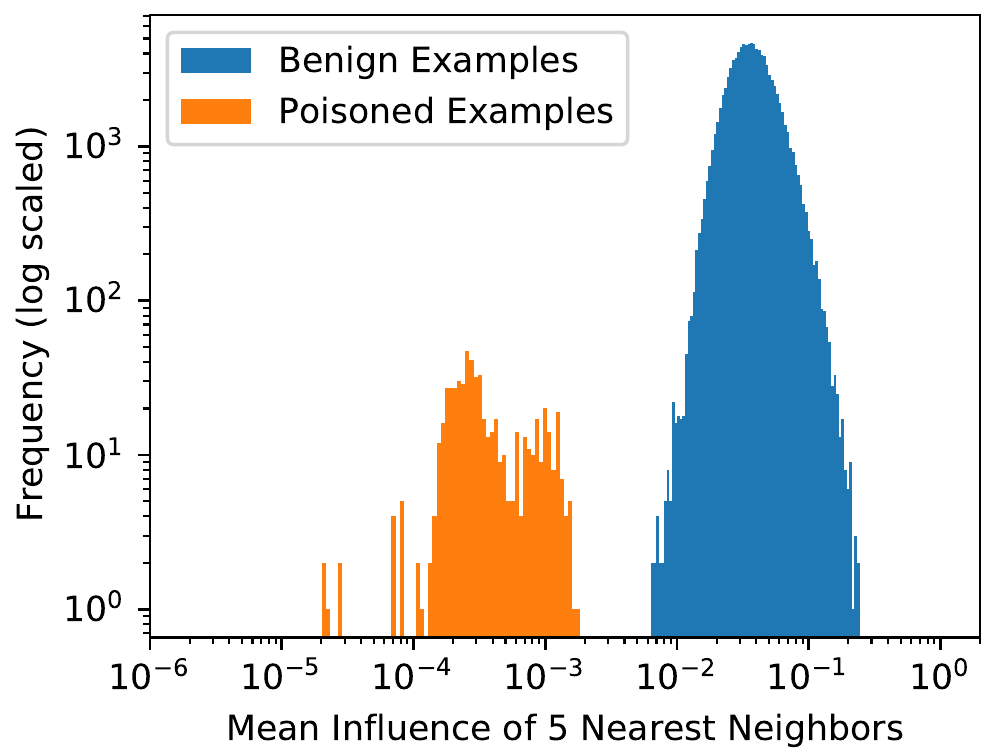}
  \caption{Our defense reliably detects poisoning attacks using MixMatch on SVHN.
  In all cases, we perfectly separate the standard training data from injected
  poisoned examples.}
  \label{fig:defensex1}
\end{figure}

\begin{figure}[H]
\centering
  \includegraphics[scale=.4]{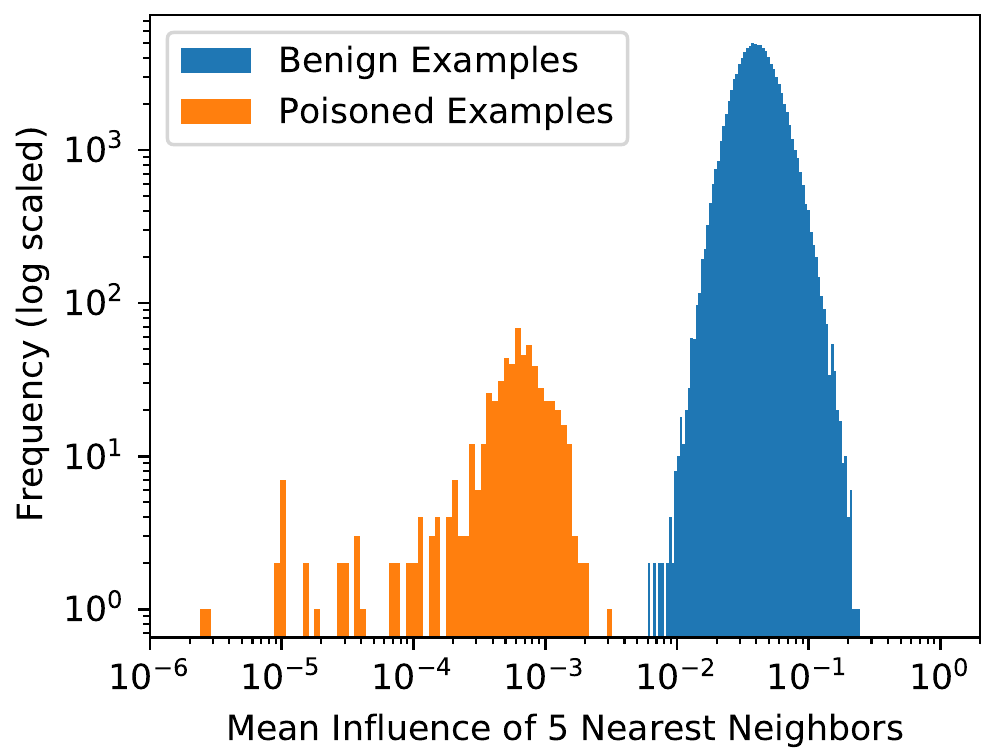}
  \includegraphics[scale=.4]{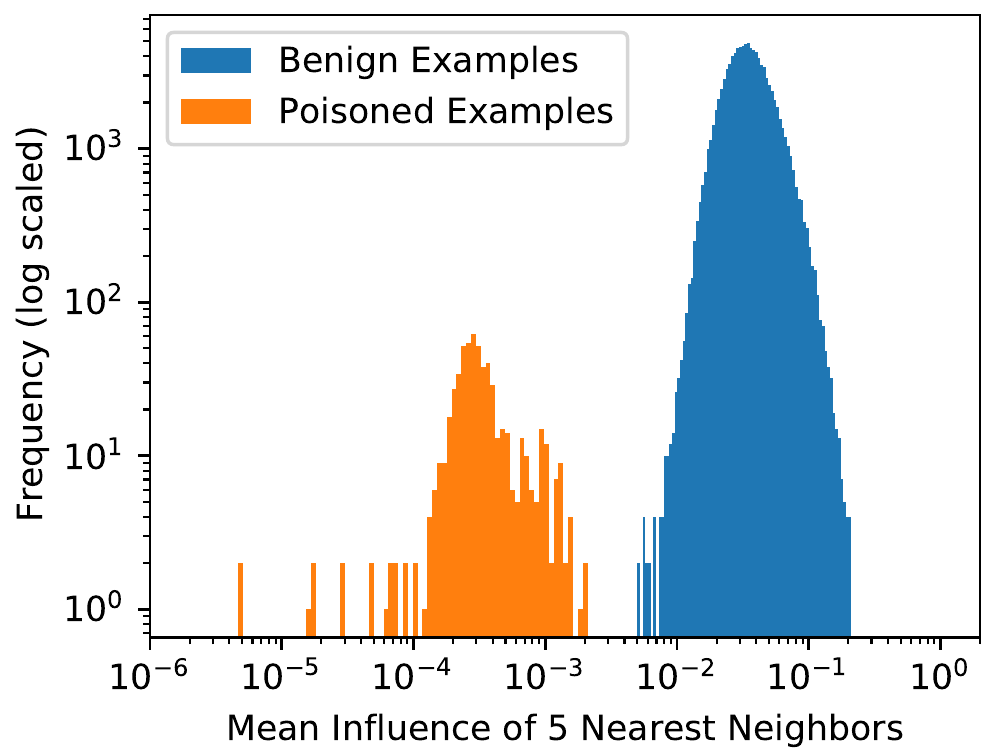}
  \caption{Our defense reliably detects poisoning attacks using FixMatch on SVHN.
  In all cases, we perfectly separate the standard training data from the injected
  poisoned examples.}
  \label{fig:defensex1}
\end{figure}

\end{document}